\documentclass[11pt]{article}
\usepackage{coling2020}
\usepackage{times}
\usepackage{url}
\usepackage{latexsym}
\usepackage{graphicx}
\usepackage{adjustbox}

\colingfinalcopy 

\title{WikiBERT models: deep transfer learning for many languages} 

\author{Sampo Pyysalo \hspace{0.5cm} Jenna Kanerva \hspace{0.5cm} Antti Virtanen \hspace{0.5cm} Filip Ginter \\
  TurkuNLP group, Department of Future Technologies\\
  University of Turku, Finland\\
  \texttt{first.last@utu.fi}
}

\date{}

\begin{document}
\maketitle
\begin{abstract}
Deep neural language models such as BERT have enabled substantial recent advances in many natural language processing tasks. Due to the effort and computational cost involved in their pre-training, language-specific models are typically introduced only for a small number of high-resource languages such as English. While multilingual models covering large numbers of languages are available, recent work suggests monolingual training can produce better models, and our understanding of the tradeoffs between mono- and multilingual training is incomplete.
In this paper, we introduce a simple, fully automated pipeline for creating language-specific BERT models from Wikipedia data and introduce 42 new such models, most for languages up to now lacking dedicated deep neural language models. We assess the merits of these models using the state-of-the-art UDify parser on Universal Dependencies data, contrasting performance with results using the multilingual BERT model. We find that UDify using WikiBERT models outperforms the parser using mBERT on average, with the language-specific models showing substantially improved performance for some languages, yet limited improvement or a decrease in performance for others. We also present preliminary results as first steps toward an understanding of the conditions under which language-specific models are most beneficial.
All of the methods and models introduced in this work are available under open licenses from \url{https://github.com/turkunlp/wikibert}.
\end{abstract}

%
    %
    %
    %
    %
    %
    %

\section{Introduction}
\label{intro}

Transfer learning using language models pre-trained on large unannotated corpora has allowed for substantial recent advances at a broad range of natural language processing (NLP) tasks. By contrast to earlier context-independent approaches such as word2vec \cite{mikolov2013efficient} and GloVe \cite{pennington2014glove}, models such as ULMFiT \cite{howard2018universal}, ELMo \cite{peters2018deep}, GPT \cite{radford2018improving} and BERT \cite{devlin2018bert} create contextualized representations of meaning, capable of providing both contextualized word embeddings as well as embeddings for longer text segments than words. Recent pre-trained language models has been rapidly advancing the state of the art in a range of natural language understanding tasks \cite{wang2018glue,wang2019superglue} as well as established NLP tasks such as named entity recognition and syntactic analysis \cite{martin2020camembert,virtanen2019multilingual}.

The transformer architecture \cite{vaswani2017attention} and the BERT language model of \newcite{devlin2018bert} have been particularly influential, with transformer-based models in general and BERT in particular fuelling a broad range of advances in natural language processing tasks over the recent years. However, most recent work introducing new deep neural language models has focused on English, with models for other languages released later, if at all. For BERT, the original study introducing the model \cite{devlin2018bert} addressed only English, and Google later released a Chinese model as well as a multilingual model, mBERT,\footnote{\url{https://github.com/google-research/bert/blob/master/multilingual.md}. The document explicitly states ``We do not plan to release more single-language models''} trained on text from 104 languages.

A range of language-specific BERT models have since been created by various groups, for example
BERTje\footnote{\url{https://github.com/wietsedv/bertje}} \cite{de2019bertje},
CamemBERT\footnote{\url{https://camembert-model.fr/}} \cite{martin2020camembert},
FinBERT\footnote{\url{https://turkunlp.org/FinBERT/}} \cite{virtanen2019multilingual}, and
RuBERT\footnote{\url{https://github.com/deepmipt/deeppavlov/}} \cite{kuratov2019adaptation}, demonstrating substantial improvements over the multilingual model in various language-specific downstream task evaluations. However, these efforts have so far not added up to a broad-coverage collection of consistent-quality language-specific deep transfer learning models, and we are not aware of previous efforts to introduce readily executable pipelines for creating data for pre-training deep transfer learning models. Here, we take steps towards addressing these issues by introducing both a simple, fully automated pipeline for creating language-specific BERT models from Wikipedia data as well as 42 new such models.

\section{Data}

We next introduce the sources of unannotated data used for pre-training and annotations used for prepreprocessing and evaluation in our work.

\subsection{Pre-training data}

The English Wikipedia was the main source of text for pre-training the original English BERT models, accounting for three-fourths of its pre-training data.\footnote{The remaining quarter of BERT pre-training data was drawn from the BooksCorpus, a unique (and now unavailable) resource for which analogous corpora in other languages are very challenging to create.}
The multilingual BERT models were likewise trained on Wikipedia data.
To roughly mirror the original BERT pre-training data selection, we chose to pre-train our models exclusively on Wikipedias in various languages.

As of this writing, the List of Wikipedias\footnote{\url{https://en.wikipedia.org/wiki/List_of_Wikipedias}} identifies Wikipedias in 309 languages. Their sizes vary widely: while the largest of the set, the English Wikipedia, contains over six million articles, the smaller half of Wikipedias (155 languages) put together only total approximately 400,000 articles.
As the BERT base model has over 100 million parameters and BERT models are frequently trained on billions of words of unannotated text, it seems safe to estimate that attempting to train BERT for e.g.\ Old Church Slavonic, ranked 272nd with fewer than 1000 articles (under 50,000 tokens), would likely not produce a very successful model. It is nevertheless not well established how much unannotated text is required to pre-train a language-specific model, and how much the domain and quality of the pre-training data affect the model performance. An evaluation carried out by \newcite{raffel2019exploring} on controlling the size and text sources of the English pre-training dataset suggests that a larger pre-training dataset does not always yield better performance on downstream tasks, and even though the pure Wikipedia data rarely achieves state-of-the-art downstream performance, it gives a competitive baseline performance. However, as previously stated one must keep in mind here that the English Wikipedia is considerably larger than Wikipedias for most other languages.

In order to focus our computational resources as well as best support the community, we have so far opted to exclude dead languages, i.e.\ languages that are not in everyday spoken use by any community, from our model pre-training pipeline. We have thus not created models for Ancient Greek, Coptic, Gothic, Latin, Old Church Slavonic, and Old French.
Other than this exclusion, we have broadly proceeded to introduce preprocessing support and models for languages in decreasing order of the size of their Wikipedias an support in Universal Dependencies, discussed below.

\subsection{Universal Dependencies}

The Universal Dependencies (UD) is a community lead effort seeking to create cross-linguistically consistent treebank annotations for many typologically different languages. \cite{nivre2016universal} As of this writing, the latest release of the UD treebanks\footnote{\url{https://universaldependencies.org/}} is v2.6, which includes 163 treebanks covering 92 languages. To maintain comparability with recent work on UD parsing, most importantly the study introducing the UDify parser \cite{kondratyuk2019}, we here use the UD v2.3 treebanks\footnote{\url{http://hdl.handle.net/11234/1-2895}}, with 129 treebanks in 76 languages.

When assessing the WikiBERT models, we limit our evaluation to the subset of UD~v2.3 treebanks that have training, development, and test sets, thus excluding e.g.\ the 17 parallel UD treebanks which only provide test sets.
We further exclude from evaluation treebanks released without text, namely \texttt{ar\_nyuad}, \texttt{fr\_ftb}, \texttt{ja\_bccwj}
as well as the Swedish sign language treebank \texttt{swl\_sslc}. Finally, we exclude \texttt{mr\_ufal} \texttt{mt\_mudt}, \texttt{te\_mtg}
and \texttt{ug\_udt} as we currently do not have dedicated BERT models for these languages.

\section{Methods}

We next briefly introduce the primary steps of the preprocessing pipeline for creating pre-training examples from Wikipedia source as well as the tools used for text processing, model pre-training, and evaluation.

\subsection{Preprocessing pipeline}

In order to create good quality data from raw Wikipedia dumps in the format required by BERT model training, we introduce a pipeline that performs the following primary steps:

\paragraph{Data and model download} The full Wikipedia database backup dump is downloaded from a mirror site\footnote{\url{https://dumps.wikimedia.org/}} and a UDPipe model for the language from the LINDAT/CLARIN repository.\footnote{\url{http://hdl.handle.net/11234/1-3131}}

\paragraph{Plain text extraction} WikiExtractor\footnote{\url{https://github.com/attardi/wikiextractor}} is used to extract plain text with document boundaries from the Wikipedia XML dump.

\paragraph{Segmentation and tokenization} UDPipe is used with the downloaded model to segment sentences and tokenize the plain text, producing text with document, sentence, and word boundaries.

\paragraph{Document filtering} A set of heuristic rules and statistical language detection\footnote{\url{https://github.com/shuyo/language-detection}} are applied to optionally filter documents based on configurable criteria.

\paragraph{Sampling and basic tokenization} A sample of sentences is tokenized using BERT basic tokenization to produce examples for vocabulary generation that match BERT tokenization criteria.

\paragraph{Vocabulary generation} A subword vocabulary is generated using the SentencePiece\footnote{\url{https://github.com/google/sentencepiece}} \cite{kudo2018sentencepiece} implementation of byte-pair encoding \cite{gage1994new,sennrich2015neural}. After generation the vocabulary is converted to the BERT WordPiece format.

\paragraph{Example generation} Masked language modeling and next sentence prediction examples using the full BERT tokenization specified by the generated vocabulary are created in the TensorFlow TFRecord format using BERT tools.

\vspace{0.5cm}
\noindent
The created vocabulary and pre-training examples can be used directly with the original BERT implementation to train new language-specific models.

\subsection{UDPipe}

UDPipe \cite{straka2016udpipe} is a parser capable of producing segmentation, part-of-speech and morphological tags, lemmas and dependency trees. In this work we use UDPipe for sentence segmentation and tokenization. The segmentation component in UDPipe is a character-level bidirectional GRU network simultaneously predicting the end-of-token and end-of-sentence markers.

\subsection{Pre-training}

We aimed to largely mirror the original BERT process in our selection of parameters and setting for the pre-training process to create the WikiBERT models, with some adjustments made to account for differences in computational resources. Specifically, while the original BERT models were trained on TPUs, we trained on Nvidia Volta V100 GPUs with 32GB memory. We followed the original BERT processing in training for a total of 1M steps in two stages, the first 900K steps with a maximum sequence length of 128, and the last 100K steps with a maximum of 512. Due to memory limitations, each model was trained on 4 GPUs using a batch size of 140 during the 128 sequence length phase, and 8 GPUs with a batch size of 20 during the 512 phase.

\subsection{UDify}

To evaluate the models, we apply the UDify parser \cite{kondratyuk2019} trained on Universal Dependencies data. UDify is a state-of-the-art model and can predict UD part-of-speech tags, morphological features, lemmas, and dependency trees, allowing several aspects of the models' capabilities to be assessed straightforwardly. UDify implements a multi-task learning objective using task-specific prediction layers on top of a pre-trained BERT encoder. All prediction layers are trained simultaneously, while also fine-tuning the pre-trained encoder weights. In the following evaluation, we focus on the parsin performance using the standard Labeled Attachment Score (LAS) metric.

\section{Results}

\renewcommand{\tabcolsep}{6pt}
\begin{table}[]
\centering
\begin{adjustbox}{max width=\textwidth}

\begin{tabular}{l|rrl}
         &        & \multicolumn{2}{c}{Average LAS} \\
Language (code) & Tokens & mBERT & WikiBERT \\ \hline
Afrikaans (af) & 24M & \textbf{87.85} & 87.33 \\ 
Arabic (ar) & 184M & 83.81 & \textbf{85.47} \\ 
Belarusian (be) & 34M & \textbf{81.77} & 79.81 \\ 
Bulgarian (bg) & 71M & 92.30 & \textbf{92.51} \\ 
Catalan (ca) & 236M & \textbf{92.08} & 92.06 \\ 
Czech (cs) & 143M & 90.45 & \textbf{90.69} \\ 
Danish (da) & 65M & 85.78 & \textbf{85.84} \\ 
German (de) & 1.0B & 83.16 & \textbf{84.13} \\ 
Greek (el) & 81M & 91.63 & \textbf{92.35} \\ 
English (en) & 2.7B & \textbf{88.09} & 88.05 \\ 
Spanish (es) & 678M & \textbf{90.42} & 90.12 \\ 
Estonian (et) & 38M & 85.86 & \textbf{87.43} \\ 
Basque (eu) & 45M & 82.99 & \textbf{83.70} \\ 
Persian (fa) & 95M & 86.60 & \textbf{88.60} \\ 
Finnish (fi) & 97M & 87.64 & \textbf{90.81} \\ 
French (fr) & 858M & \textbf{89.22} & 88.77 \\ 
Galician (gl) & 58M & \textbf{83.05} & 82.61 \\ 
Hebrew (he) & 166M & 88.77 & \textbf{90.17} \\ 
Hindi (hi) & 35M & 91.59 & \textbf{91.86} \\ 
Croatian (hr) & 54M & \textbf{89.46} & 89.40 \\ 
Hungarian (hu) & 129M & 83.99 & \textbf{86.21} \\ 

\end{tabular}
\qquad
\begin{tabular}{l|rrl}
         &        & \multicolumn{2}{c}{Average LAS} \\
Language (code) & Tokens & mBERT & WikiBERT \\ \hline
Indonesian (id) & 93M & \textbf{80.40} & 80.12 \\ 
Italian (it) & 579M & 89.64 & \textbf{89.77} \\ 
Japanese (ja) & 596M & 92.78 & \textbf{92.92} \\ 
Korean (ko) & 79M & 86.19 & \textbf{87.28} \\ 
Lithuanian (lt) & 34M & \textbf{58.68} & 58.40 \\ 
Latvian (lv) & 21M & 84.29 & \textbf{84.46} \\ 
Dutch (nl) & 300M & 90.26 & \textbf{91.02} \\ 
Norwegian (no) & 112M & 91.54 & \textbf{91.94} \\ 
Polish (pl) & 282M & 94.45 & \textbf{95.58} \\ 
Portuguese (pt) & 326M & 91.91 & \textbf{92.21} \\ 
Romanian (ro) & 85M & \textbf{86.83} & 86.52 \\ 
Russian (ru) & 565M & 90.35 & \textbf{91.13} \\ 
Slovak (sk) & 39M & 91.64 & \textbf{91.73} \\ 
Slovenian (sl) & 42M & 92.83 & \textbf{93.37} \\ 
Serbian (sr) & 96M & \textbf{92.30} & 91.79 \\ 
Swedish (sv) & 364M & 86.42 & \textbf{87.12} \\ 
Tamil (ta) & 26M & \textbf{70.14} & 69.63 \\ 
Turkish (tr) & 71M & 69.33 & \textbf{71.25} \\ 
Ukrainian (uk) & 260M & 88.57 & \textbf{90.41} \\ 
Urdu (ur) & 18M & \textbf{82.66} & 82.15 \\ 
Vietnamese (vi) & 172M & 66.89 & \textbf{68.87} \\ 
\end{tabular}
\end{adjustbox}
\caption{Summary of Wikipedia training data size (Tokens) and average LAS results for UDify for Universal Dependencies treebanks in each language with mBERT and WikiBERT initialization. }
\label{tab:results}
\end{table}

\begin{figure}
\centering
\includegraphics[width=\textwidth]{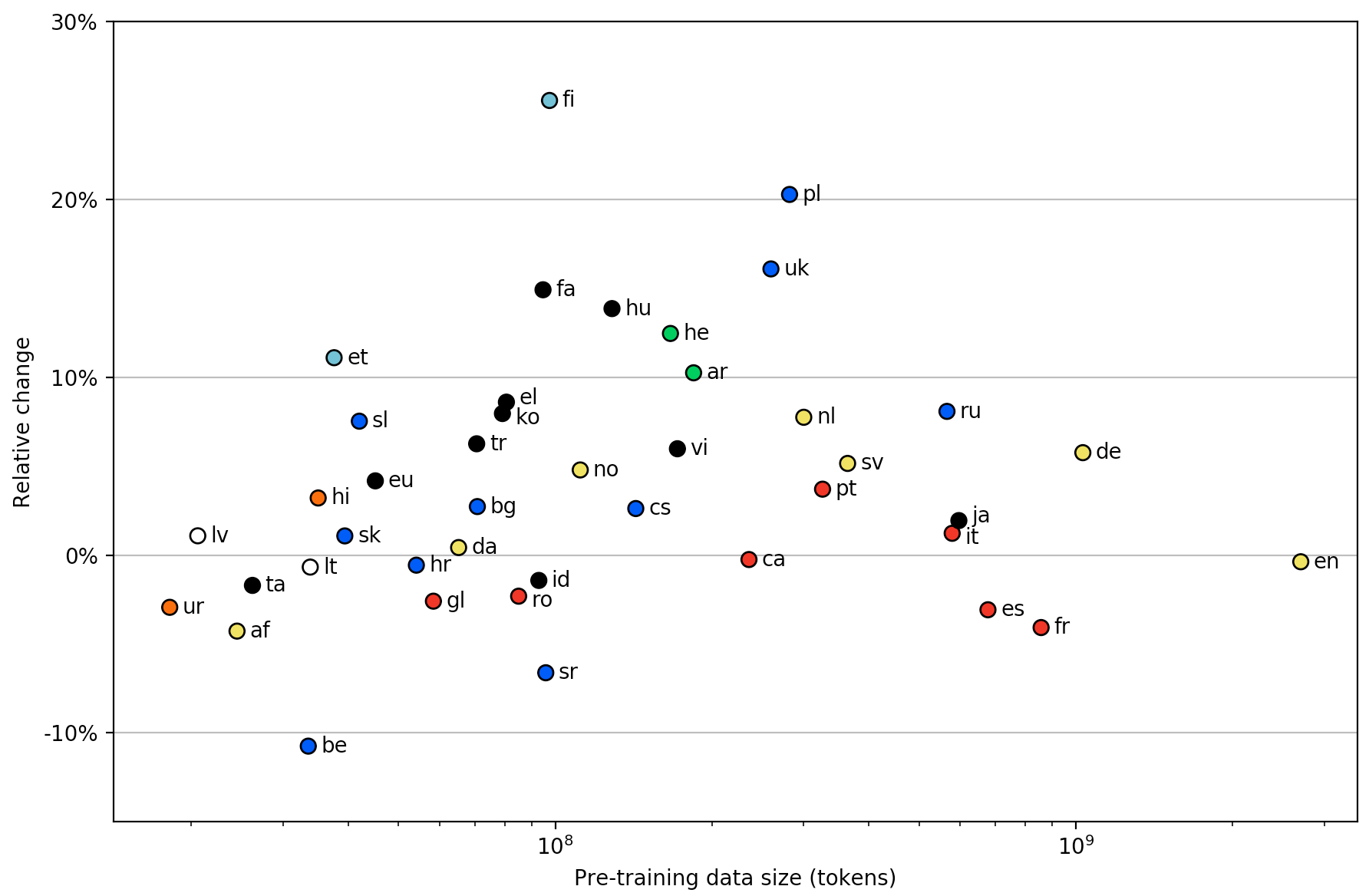}
\caption{Average relative change in LAS when replacing mBERT with a WikiBERT model for UDify initialization plotted against the WikiBERT pre-training data size in tokens. Coloring indicates language grouping by genera (black = other).}
\label{fig:las-comparison}
\end{figure}

Table~\ref{tab:results} summarizes the evaluation results. We find a complex, mixed picture where mBERT and WikiBERT models each appear clearly superior for different languages, for example, mBERT for Belarusian and WikiBERT for Finnish. On average across the languages, UDify initialized with WikiBERT models slightly edges out mBERT initialization, with 86.1\% average for mBERT and 86.6\% for WikiBERT (an approximately 4\% relative decrease in LAS error). However, such averaging hides more than it reveals, and it is much more interesting to consider the various potential impacts on performance from pre-training data size, potential support from close relatives in the same language family, and other similar factors. The various UD treebanks represent very different levels of challenge, with LAS results ranging from below 60\% to above 95\%. To reduce the impact of the properties of the treebanks on the comparison, in the following we focus on the relative change in performance when initializing UDify with a WikiBERT model compared to the baseline approach using mBERT.

Figure~\ref{fig:las-comparison} shows the average relative change in performance over all treebanks for a language when replacing mBERT with the relevant WikiBERT model for UDify, plotted against the number of tokens in Wikipedia for the language. While the data is very noisy due to a number of factors, we find some indication of a ``sweet spot'' where training a dedicated language models tends to show most benefit over using the multilingual model when at least approximately 100M tokens but fewer than 1B tokens of pre-training data are available. We also briefly note some other properties in this data:
\begin{itemize}
\item For English, a language in the large Germanic family and the language with the largest amount of pre-training data, mBERT and WikiBERT results are effectively identical.
\item The greatest loss when moving from mBERT to a WikiBERT model is seen for Belarusian, a slavic language closely related to Russian, for which considerably more training data is available.
\item The greatest gain when moving from mBERT to a WikiBERT model is seen for Finnish, a comparatively isolated language that nevertheless has a reasonably-sized Wikipedia.
\end{itemize}
Observations such as these may suggest fruitful avenues for further research into the conditions under which mono- and multilingual language model training is expected to be most successful.

\section{Discussion}

This short manuscript has provided a first brief introduction to the WikiBERT models, a collection of dedicated language-specific BERT models covering many languages that previously lacked a dedicated deep transfer learning model of this type. We demonstrated the value of these models compared to the multilingual BERT model through evaluation on the Universal Dependencies multilingual dependency parsing data, showing that a WikiBERT model will provide better performance than multilingual BERT on average, and in multiple cases providing a more than 10\% relative decrease in LAS error compared to the multilingual model.

The availability of the WikiBERT collection of models opens up a broad range of potential avenues for research into the strengths, weaknesses and challenges in both mono- and multilingual language modeling. Due to scheduling constraints, this initial manuscript must necessarily leave most such questions for future work.

\section*{Acknowledgements}

We gratefully acknowledge the support of the Academy of Finland, and CSC --- the Finnish IT Center for Science for providing computational resources for this effort.

\bibliographystyle{coling}
\bibliography{main}

\end{document}